%% file: main.tex
\documentclass{article}

\usepackage[preprint]{neurips_2025}

\usepackage[utf8]{inputenc} % allow utf-8 input
\usepackage[T1]{fontenc}    % use 8-bit T1 fonts
\usepackage{hyperref}       % hyperlinks
\usepackage{url}            % simple URL typesetting
\usepackage{booktabs}       % professional-quality tables
\usepackage{amsfonts}       % blackboard math symbols
\usepackage{nicefrac}       % compact symbols for 1/2, etc.
\usepackage{microtype}      % microtypography
\usepackage{xcolor}         % colors

\usepackage[utf8]{inputenc}
\usepackage[T1]{fontenc}
\usepackage{microtype}
\usepackage{booktabs}
\usepackage{amsmath}
\usepackage{textcomp}
\usepackage{xcolor}
\usepackage{pgfpages}
\usepackage{pgfplots}
\usepackage{pgfplotstable}
\usepackage{tikz}
\usepackage{enumerate}
\usepackage[noend]{algpseudocode}
\usepackage{listings}
\usepackage{url}
\usepackage[multiple]{footmisc}
\usepackage{multirow}
\usepackage{arydshln}
\usepackage{xspace}
\usepackage{flushend}
\usepackage{diagbox}

\usepackage{bm}
\usepackage{siunitx}
\usepackage{multirow}
\usepackage{enumitem}
\usepackage[breakable]{tcolorbox}

\usepackage[linesnumbered,ruled,vlined]{algorithm2e}
\usepackage{graphicx}
\usepackage{subfig}
\usepackage{wrapfig}
\usepackage{balance}
\title{AR2: Attention-Guided Repair for the Robustness of CNNs Against Common Corruptions}

\author{%
  Fuyuan Zhang\thanks{Corresponding author: Fuyuan Zhang}\\
  Zhejiang University, China\\
  \texttt{fuyuanzhang@163.com}\\
  \And
  Qichen Wang\\
  Kyushu University, Japan\\
  \texttt{wang.qichen.256@s.kyushu-u.ac.jp}
  \And
  Jianjun Zhao\\
  Kyushu University, Japan\\
  \texttt{zhao@ait.kyushu-u.ac.jp}
  }

\begin{document}

\maketitle

\begin{abstract}
  Deep neural networks suffer from significant performance degradation when exposed to common corruptions such as noise, blur, weather, and digital distortions, limiting their reliability in real-world applications. In this paper, we propose \emph{AR2} (Attention-Guided Repair for Robustness), a simple yet effective method to enhance the corruption robustness of pretrained CNNs. AR2 operates by explicitly aligning the class activation maps (CAMs) between clean and corrupted images, encouraging the model to maintain consistent attention even under input perturbations. Our approach follows an iterative repair strategy that alternates between CAM-guided refinement and standard fine-tuning, without requiring architectural changes. Extensive experiments show that AR2 consistently outperforms existing state-of-the-art methods in restoring robustness on standard corruption benchmarks (CIFAR-10-C, CIFAR-100-C and ImageNet-C), achieving a favorable balance between accuracy on clean data and corruption robustness. These results demonstrate that AR2 provides a robust and scalable solution for enhancing model reliability in real-world environments with diverse corruptions.

\end{abstract}

%------------------------------------------------
\section{Introduction}\label{sec:introduction}
\input{introduction.tex}

%------------------------------------------------
\section{Motivation and Overview}\label{sec:motivation}
\input{motivation.tex}

%------------------------------------------------
\section{Methodology}\label{sec:method}
\input{methodology.tex}

%------------------------------------------------
\section{Experimental Results}\label{sec:experiment}
\input{experiment.tex}
%------------------------------------------------
\section{Related Works}\label{sec:related}
\input{related-work.tex}

%------------------------------------------------
\section{Conclusion}\label{sec:conclusion}
\input{conclusion.tex}
%------------------------------------------------
\bibliographystyle{plainnat}
\bibliography{reference}

\end{document}

%% file: introduction.tex
Convolutional neural networks (CNNs) have achieved remarkable success in image classification and visual recognition tasks, yet they remain surprisingly vulnerable to common corruptions such as noise, blur, weather, and digital distortions~\cite{hendrycks2019robustness}. Even moderate perturbations, frequent in real-world environments, can severely degrade CNN performance. This sensitivity undermines their reliability for safety-critical applications or real-world deployment scenarios.

A number of approaches have been developed to improve the robustness of neural networks against common corruptions~\cite{hendrycks2020augmix,GaoSaha2020,YuQi2022}. Some methods focus on data augmentation, such as AugMix~\cite{hendrycks2020augmix}, which leverages auxiliary data to enhance generalization under distribution shifts. Others, like Sensei~\cite{GaoSaha2020}, employ mutation-based fuzzing techniques to generate augmented training data, systematically exposing and correcting model vulnerabilities through genetic search optimization. Nevertheless, a key challenge remains: achieving an acceptable balance between accuracy on clean data and robustness against corrupted inputs. Consequently, models must simultaneously maintain high accuracy on both clean and corrupted images—two inherently mismatched distributions—presenting a fundamentally difficult task. Despite progress, robustness against common corruptions remains an open challenge.

We propose AR2, a two-stage approach designed to improve the robustness of convolutional neural networks against common corruptions through attention-guided repair. In the first stage, called CAM-guided refinement, AR2 leverages class activation maps (CAMs)~\cite{ZhouKhosla2016} to estimate the attention of CNNs on both clean and corrupted versions of the same image. It then guides the network to align these attention maps, encouraging consistent focus across clean and corrupted inputs. This alignment helps the model maintain reliable predictions despite input degradations. In the second stage, AR2 applies a fine-tuning process to mitigate any potential drop in accuracy, further enhancing performance on both clean and corrupted images. Our approach requires only the original training dataset (e.g., ImageNet) and its synthetically corrupted versions (generated using standard corruption functions), needing neither additional external datasets nor architectural modifications.

We evaluate AR2 against three state-of-the-art techniques for improving the robustness of deep neural networks against common corruptions. Our evaluation uses CIFAR-10-C, CIFAR-100-C and ImageNet-C~\cite{hendrycks2019robustness}, the standard corrupted variants of CIFAR-10/100~\cite{Krizhevsky2008} and ImageNet~\cite{RussakovskyDeng15}, covering all 15 corruption types across both benchmarks. Experimental results demonstrate that AR2 consistently improves corruption robustness, outperforming existing methods by a significant margin. These results validate AR2 as an effective general framework for enhancing CNN robustness across diverse corruption scenarios, from noise and blur to weather and digital distortions.

\textbf{Contributions.} We make the following contributions:
\begin{enumerate}
\item We propose AR2, a simple yet effective attention-guided repair approach for improving the robustness of convolutional neural networks against common corruptions.

\item Through comprehensive evaluation on CIFAR-10-C, CIFAR-100-C and ImageNet-C across all 15 corruption types, we demonstrate that AR2 consistently outperforms state-of-the-art robustness enhancement techniques.

\item We show through experiments that AR2's targeted repairs for specific corruptions rarely compromise robustness against other corruption types, demonstrating its ability to address vulnerabilities without introducing harmful side effects.
\end{enumerate}

\textbf{Outline.} Section~\ref{sec:motivation} presents the motivation behind our work and provides an overview of the proposed approach. In Section~\ref{sec:method}, we introduce AR2, our attention-guided repair method for improving CNN robustness against common corruptions. Section~\ref{sec:experiment} reports experimental results demonstrating the effectiveness of AR2. We discuss related work in Section~\ref{sec:related} and conclude the paper in Section~\ref{sec:conclusion}.

%% file: motivation.tex
\subsection{Motivations}
Convolutional neural networks have achieved remarkable success across a wide range of applications. However, their black-box nature makes it difficult to understand how they make decisions, which in turn complicates efforts to ensure critical properties such as robustness, fairness, and reliability—properties essential for safe deployment in real-world scenarios. To address this, considerable research has focused on improving the interpretability of CNNs. In particular, techniques such as saliency maps~\cite{ZhouKhosla2016,SelvarajuCogswell2017,WangWang2020,DanielNikhil2017,ChattopadhaySarkar2018,SundararajanTaly2017} have been developed to visually illustrate where a network "attends" in an input image during prediction, often using heatmaps to highlight areas of high and low attention.

% fig 1-------------------------------------begin---------------------------------
\begin{figure}
\includegraphics[width=\linewidth]{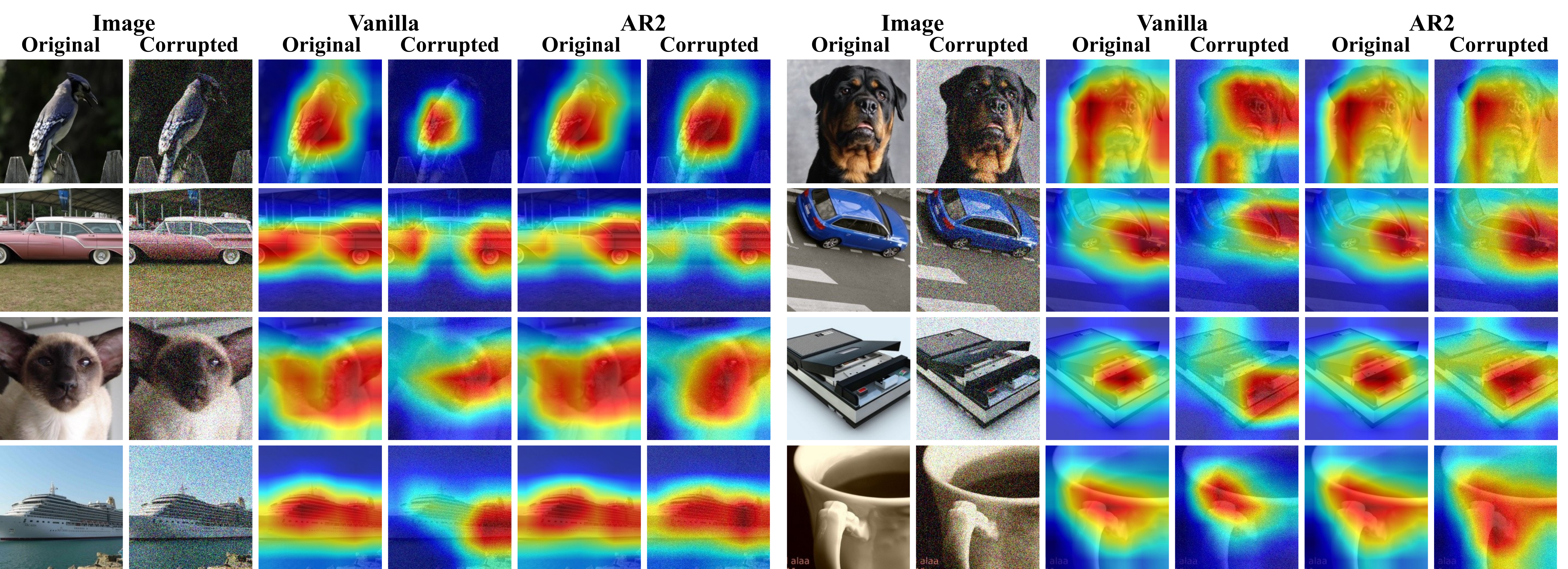}
\caption{Attention shifts under corruption and AR2's corrective effects. Eight example groups, each showing (1) original image, (2) corrupted image, (3-4) CAMs from the vanilla network (naturally trained network), and (5-6) CAMs from the AR2-repaired network. Comparisons demonstrate how corruptions disrupt attention localization in vanilla models and how AR2 mitigates these shifts.}
\label{fig:image_grid}
\end{figure}
% fig 1-------------------------------------end-----------------------------------

We leverage class activation maps (CAMs)~\cite{ZhouKhosla2016} to study how corruptions affect CNN attention mechanisms. Our analysis demonstrates that corruptions systematically disrupt feature localization in vanilla models. While vanilla models produce focused CAM activations on class-discriminative regions for clean images, their responses become shifted or fragmented when processing corrupted versions (Figure~\ref{fig:image_grid}). These misalignments strongly correlate with accuracy degradation, confirming that corruptions both mask essential features and introduce misleading signals that steer the model toward incorrect interpretations.

These insights motivate our proposed solution: to enhance CNN robustness by correcting their attention when processing corrupted inputs. We hypothesize that consistent, appropriate attention across clean and corrupted images will yield more reliable predictions. Based on this hypothesis, we developed AR2, an attention-guided repair approach that aligns the network's attention across input conditions. As Figure~\ref{fig:image_grid} demonstrates, AR2-repaired networks maintain stable focus on class-discriminative regions despite corruptions, unlike the erratic shifts in vanilla models. Our experiments confirm these improvements translate to significant robust accuracy gains (Section 4). 

\subsection{The Workflow of AR2}
The AR2 framework consists of two sequential steps: \emph{CAM-guided refinement} and \emph{fine-tuning}, each serving a distinct purpose in improving the robustness of CNNs against common corruptions.

\begin{figure}[hbp]
  \centering
  \includegraphics[width=0.9\linewidth]{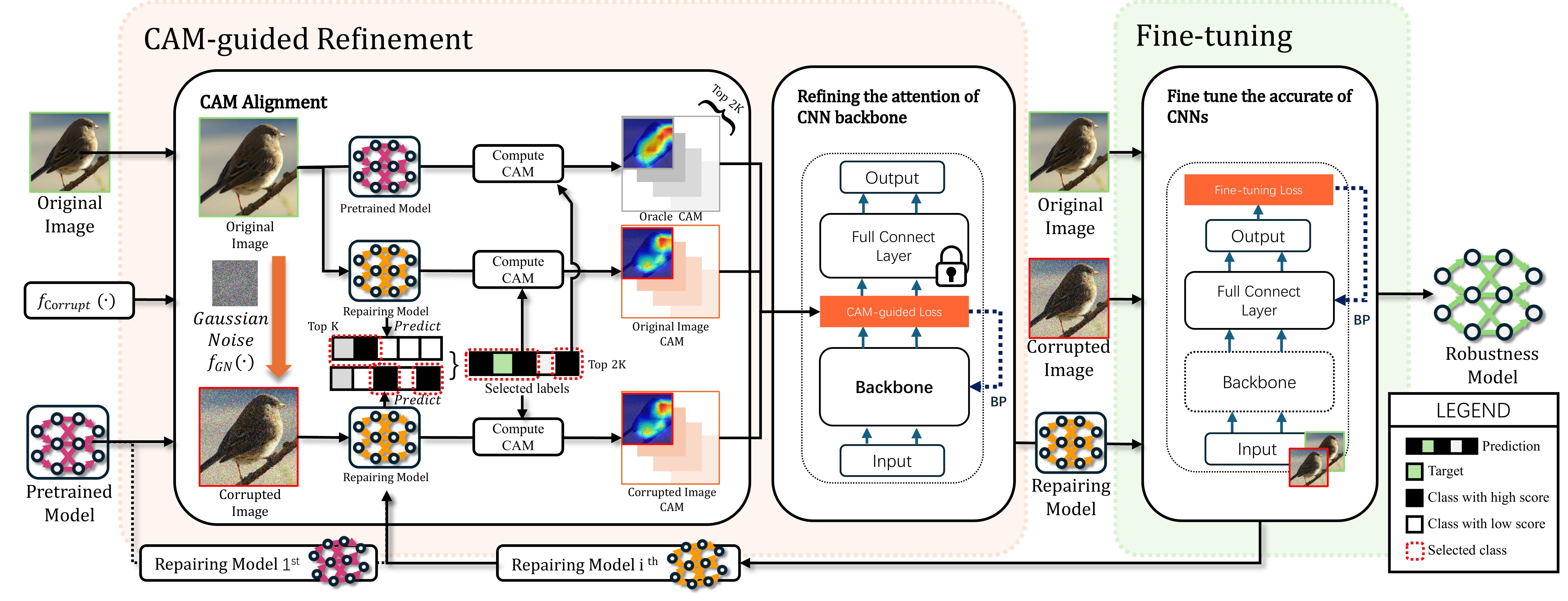}
  \caption{Workflow of AR2. Our approach consists of CAM-guided refinement and fine-tuning to improve robustness against corrupted images.}
  \label{fig:ar2_workflow}
\end{figure}

In the CAM-guided refinement step, we aim to align the model’s attention maps between clean and corrupted images. Specifically, we encourage the network to attend to similar regions in both clean and corrupted versions of the same image, using class activation maps as the attention signal. The motivation is that corrupted images often lead to shifted or degraded attention, which contributes to incorrect predictions. By explicitly guiding the model to preserve its attention focus under corruption, we expect to enhance robustness. However, this stage does not directly optimize classification accuracy; its goal is purely to enforce attention consistency, which is necessary—but not sufficient—for good performance.

In the fine-tuning step, we address the accuracy of the model. Since CAM-guided refinement alone may cause accuracy to degrade (due to the lack of a classification objective), we follow up with supervised fine-tuning to restore or improve performance on both clean and corrupted images. Importantly, these two stages are performed separately, without any joint training objective.

To support the CAM-guided refinement process, AR2 maintains two models during training:
\begin{enumerate}
\item The reference model, which is the original naturally trained network. This model remains frozen throughout refinement and provides stable CAMs for clean images.

\item The repairing model, which is a copy of the reference model at initialization. This model is updated during AR2 and learns to align its attention on corrupted images with that of the reference model on clean images.
\end{enumerate}

At each repairing iteration, we generate corrupted versions of the training images and use both models to compute CAMs. The CAM alignment loss is then used to update the repairing model. This separation ensures that the clean image attention remains stable and reliable, while the repairing model gradually adapts to handle corrupted inputs more robustly.

%% file: methodology.tex
\subsection{Class Activation Maps}
Class activation maps (CAMs)~\cite{ZhouKhosla2016} provide visual explanations by highlighting discriminative image regions that contribute to a model's prediction.  Given an input image $x$, let $f(x; \theta) \in \mathbb{R}^{H \times W \times C}$ denote the feature map from the last convolutional layer, where each channel $f_k(x; \theta)$ is a spatial map activated by the $k$-th channel. For a target class $c$, the CAM is computed as:
\begin{equation}
M_c(x; \theta) = \sum_{k=1}^{C} w_k^c \cdot f_k(x; \theta),
\end{equation}
where $w_k^c$ is the weight connecting the $k$-th channel to class $c$ in the fully connected layer. The CAM $M_c(x; \theta)$ is upsampled to the input resolution and normalized for visualization. While CAMs are traditionally used for interpretability, we extend their role as supervisory signals for improving model robustness. In the following section, we present our CAM-guided refinement method, which aligns CAMs of clean and corrupted inputs to achieve more robust feature representations.

\subsection{CAM-Guided Refinement}
\label{sec:cam_refinement}
The goal of the CAM-guided refinement step is to improve the model's robustness against corrupted images by aligning the class activation maps between clean and corrupted inputs. By encouraging the model to treat clean and corrupted images similarly at the feature level, we aim to enhance robust accuracy while carefully controlling the accuracy drop on clean data.

This step operates using two models. The first is a fixed \emph{reference model}, which is the original pretrained model and remains unchanged throughout refinement. Its role is to generate reliable CAMs from clean images, which serve as the alignment target. The second is the \emph{model being repaired}, which is initialized from the same pretrained weights but is iteratively updated to better align its CAMs.

Given a clean image $x$ and its corrupted counterpart $x'$, we compute CAMs from both models. The training loss minimizes the sum of two mean squared errors (MSE): one between the CAM of the clean image from the repaired model and the reference CAM, and the other between the CAM of the corrupted image and the same reference CAM.

Specifically, we select a set of classes $\mathcal{C}(x, x')$ that includes the top-$k$ predicted classes from both $x$ and $x'$, resulting in at most $2k$ classes for alignment. Formally, 
\[
\mathcal{C}(x, x') = \text{Top-}k\text{-classes}(x) \cup \text{Top-}k\text{-classes}(x').
\]
The loss function is defined as:
\begin{equation}\label{eq:cam_loss}
\mathcal{L}_\mathrm{CAM} = \sum_{c \in \mathcal{C}(x, x')} \Big[ \text{MSE}\big(M_c(x; \theta), M_c(x; \theta_{\text{ref}})\big) + \alpha\cdot\text{MSE}\big(M_c(x'; \theta), M_c(x; \theta_{\text{ref}})\big) \Big],
\end{equation}
where $M_c(x; \theta)$ denotes the CAM for class $c$ generated by the model being repaired, $M_c(x; \theta_{\text{ref}})$ is the reference CAM obtained from the fixed pretrained model, and $\alpha$ is a parameter that balances their relative contributions. This formulation encourages CAMs from both clean and corrupted images to align with the clean reference, leading to more robust feature representations. We found this approach—aligning both clean and corrupted CAMs to a stable reference—more effective than directly matching clean and corrupted CAMs, which led to suboptimal robustness in our experiments.

The CAM-guided refinement process updates only the convolutional backbone to align clean and corrupted feature activations (using top-$2k$ classes for computational focus), while keeping feedforward layers frozen to maintain their original decision boundaries. This design isolates feature-space adaptation from classification, enabling systematic robustness improvements through targeted backbone refinement that respects the model's modular architecture.

\subsection{The AR2 Approach}\label{subsec:AR2}

We now describe the overall AR2 approach, which integrates the proposed CAM-guided refinement with standard fine-tuning in an iterative manner. AR2 aims to progressively improve the robustness of a pretrained model against image corruptions while maintaining its clean accuracy. 

The AR2 workflow operates over multiple outer iterations. In each iteration, the model undergoes two stages: (i) \emph{CAM-guided refinement}, where the CAMs of clean and corrupted images are aligned following the method described in Section~\ref{sec:cam_refinement}, and (ii) \emph{fine-tuning}, where standard training is performed to recover accuracy. Both stages are performed for multiple steps before switching; specifically, each iteration consists of $N$ steps of CAM-guided refinement followed by $M$ steps of fine-tuning. 

Throughout the process, a fixed \emph{reference model}---which is the original pretrained model---is used to compute reference CAMs and is never updated. Fine-tuning is performed using both clean and corrupted images, ensuring that the model learns to generalize under corruptions while regaining performance on clean data. 

Algorithm~\ref{alg:ar2} summarizes the AR2 approach, where each outer iteration alternates between CAM-guided refinement and fine-tuning. In our current implementation, AR2 employs a per-corruption repair paradigm, where each procedure targets a specific corruption type (e.g., Gaussian noise) using exclusively corresponding corrupted images. This approach facilitates targeted feature-space adaptations for distinct corruption families.

\begin{algorithm}[h]
\caption{The AR2 Approach}
\label{alg:ar2}
\KwIn{Pretrained model $\theta$, fixed reference model $\theta_\mathrm{ref}$, clean images $x$, corrupted images $x'$, number of outer iterations $T$, numbers $N$ (refinement steps), $M$ (fine-tuning steps)}
\For{$t = 1$ to $T$}{
    \tcp{CAM-guided refinement stage}
    \For{$n = 1$ to $N$}{
        Compute CAM alignment loss $\mathcal{L}_\mathrm{CAM}$ (Eq.~\ref{eq:cam_loss})\;
        Update model $\theta$ using gradient descent on $\mathcal{L}_\mathrm{CAM}$\;
    }
    \tcp{Fine-tuning stage}
    \For{$m = 1$ to $M$}{
        Compute classification loss (e.g., cross-entropy) on clean and corrupted images\;
        Update model $\theta$ using gradient descent\;
    }
}
\KwOut{Repaired model $\theta$}
\end{algorithm}

%% file: experiment.tex
We evaluate AR2 by applying it to enhance the robustness of CNNs against common corruptions, comparing its effectiveness with state-of-the-art robustness techniques across standard benchmarks. Consistent with AR2's per-corruption design (Section~\ref{subsec:AR2}), we repair and test each corruption type separately—a strategy that mirrors real-world needs for targeted robustness. Our experiments are designed around the following research questions:

\begin{description}
    \item[RQ1:] Is AR2 effective in improving the corruption robustness of CNNs compared to existing state-of-the-art methods?
    \item[RQ2:] Does improving robustness against one type of corruption negatively impact robustness against other types of corruptions?
    \item[RQ3:] How does AR2 affect the accuracy of CNNs on clean images?
    \item[RQ4:] How important is the CAM-guided refinement step in achieving robustness improvements?
\end{description}

We release our implementation and experimental results to encourage reproducibility and foster future work in this direction.

\subsection{Evaluation Setup}

\textbf{Datasets and network models.} We evaluate the effectiveness of AR2 and compare it with state-of-the-art methods using two widely adopted benchmarks for assessing robustness against common corruptions: CIFAR-$10$-C, CIFAR-$100$-C and ImageNet-C~\cite{hendrycks2019robustness}. These benchmarks comprise $15$ algorithmically generated corruption types grouped into four categories: noise, blur, weather, and digital distortions. Each corruption type includes five severity levels, yielding $75$ distinct corruption variants for comprehensive evaluation.
For evaluation, we adhere to the standard benchmark protocol, assessing robustness across all five severity levels of corruption. Our experiments employ ResNet-34 as the backbone architecture for CIFAR-10 and CIFAR-100, and ResNet-50 for ImageNet, following established practice in the field.

\textbf{Generation of corrupted images in AR2.} AR2 improves CNN corruption robustness by aligning attention maps between clean and corrupted inputs. To enable this alignment, we generate corrupted images during the repair process using the same corruption types and protocols as CIFAR-10-C, CIFAR-100-C and ImageNet-C, but exclusively at severity level $3$.

For CIFAR-10 and CIFAR-100, corrupted images are dynamically generated during each repair iteration, exposing the model to varied corrupted samples and preventing overfitting to specific corruption patterns. For ImageNet, computational efficiency necessitates static corruption generation—a single corrupted version of each image is pre-computed and reused throughout all repair iterations. This optimizes the trade-off between augmentation diversity and computational practicality.

\textbf{Parameter settings.} Our experimental setup uses the following parameters: For CIFAR-$10$ and CIFAR-$100$ experiments with ResNet-$34$, we employ $30$ total repair iterations ($T=30$), where each iteration consists of $1$ CAM-guided refinement step ($N=1$) followed by $1$ fine-tuning step ($M=1$), with top-$2k$ class selection using $k=3$. We set $\alpha=1$ in Eq.~\ref{eq:cam_loss}. For ImageNet experiments with ResNet-$50$, we use $10$ repair iterations ($T=10$), with each iteration containing $2$ CAM-guided refinement steps ($N=2$) followed by $1$ fine-tuning step ($M=1$), and set $k=5$ for the top-$2k$ class selection. We set $\alpha=0.8$ in Eq.~\ref{eq:cam_loss}.  This configuration ensures comprehensive repair while maintaining computational efficiency across different network scales.

\textbf{Existing approaches.} We compare AR2 with three state-of-the-art techniques for improving the robustness of CNNs against common corruptions: AugMix~\cite{hendrycks2020augmix}, a data augmentation method combining stochastic augmentations with a consistency loss; Sensei~\cite{GaoSaha2020}, a fuzzing-based data augmentation approach developed in the software engineering community; and DeepRepair~\cite{YuQi2022}, a style-guided learning method that transfers the style of noise onto clean images to augment training data.

\subsection{Experimental Results}

\begin{figure}[hbp]
    \centering
    \subfloat[CIFAR-10-C]{\includegraphics[width=0.49\textwidth]{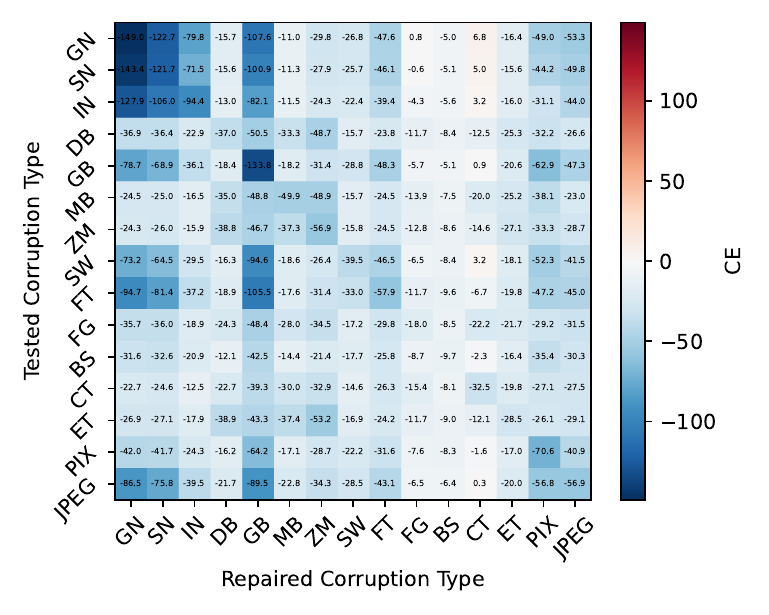}\label{fig:sub1}}
    \hfill
    \subfloat[ImageNet-C]{\includegraphics[width=0.49\textwidth]{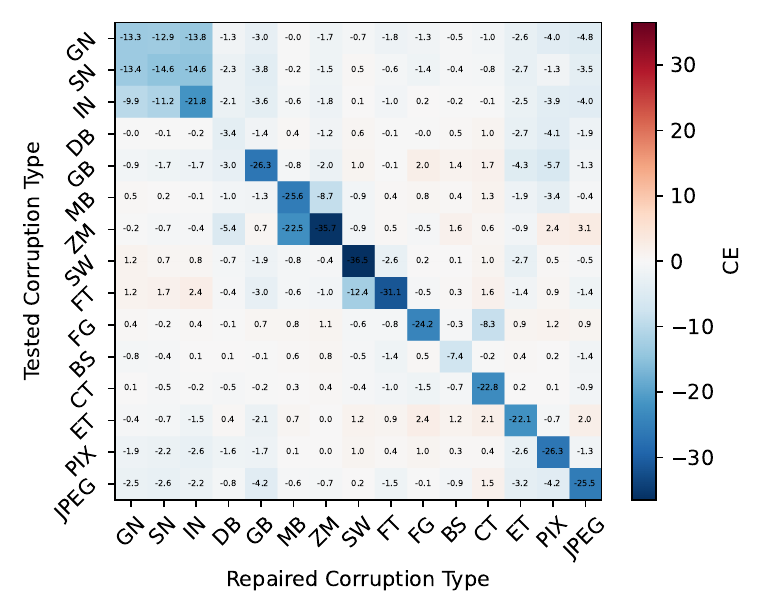}\label{fig:sub2}}
    \caption{Corruption Transfer Matrix on CIFAR-10-C and ImageNet-C. Each column shows a repaired model’s CE changes across all corruptions; each row summarizes all models’ impact on one corruption. Diagonal entries indicate each model's improvement on its target corruption.}
    \label{fig:transfer-matrix-combined}
\end{figure}

AR2 outperforms state-of-the-art methods in robustness accuracy on corrupted images while maintaining competitive clean accuracy. Experimental results are shown in Table~\ref{tab:CE} and~\ref{tab:clean_error} and Figure~\ref{fig:transfer-matrix-combined}.

\textbf{Results on effectiveness (RQ1).} Our approach demonstrates state-of-the-art performance on CIFAR-10-C, CIFAR-100-C and ImageNet-C benchmarks (Table~\ref{tab:CE}), achieving the lowest mean Corruption Error (mCE) and dominating the majority of individual corruption types. On CIFAR-10-C, we outperform all three compared methods with a $30.4\%$ mCE ($10.1\%$ better than the next-best approach). On CIFAR-100-C, we achieve state-of-the-art performance with a $48.7\%$ mCE, outperforming all baseline methods. Similarly, on ImageNet-C, we reduce the mCE by $10.1\%$ compared to AugMix, while achieving superior performance in $12$ out of $15$ corruption types.

% RQ1 -----------------------------------------------------------------
\begin{table}\scriptsize
\caption{Clean Error and Corruption Error (CE/mCE) on CIFAR-10-C, CIFAR-100-C and ImageNet-C. AR2's clean error reflects the maximum error among its 15 corruption-specific models (due to per-corruption repair), compared to baselines' single-model clean error.}
\label{tab:CE}
\setlength{\tabcolsep}{2.9pt}
\centering
\begin{tabular}{llccccccccccccccccc}

\toprule
& \multirow{2}{*}{Repair Method} &  & \multicolumn{3}{c}{Noise} & \multicolumn{4}{c}{Blur} & \multicolumn{4}{c}{Weather} & \multicolumn{4}{c}{Digital} & \\
\cmidrule(lr){4-6} \cmidrule(lr){7-10} \cmidrule(lr){11-14} \cmidrule(lr){15-18}
&   & Clean & GN & SN & IN & DB & GB & MB & ZM & SW & FT & FG & BS & CT & ET & PIX & JPEG & mCE \\
% CIFAR100 -----------------------------------------------------------------------------
\midrule
\multirow{6}{*}{\rotatebox{90}{CIFAR-10}}
 & Vanilla & 8.6 & 186.0 & 157.7 & 126.9 & 64.6 & 179.9 & 74.2 & 81.7 & 68.6 & 84.1 & 38.5 & 32.5 & 51.1 & 63.9 & 104.2 & 98.9 & 94.2\\

 & Fine-tuning & 9.2 & 166.0 & 145.7 & 126.7 & 55.3 & 148.5 & 60.3 & 62.0 & 62.7 & 64.6 & 29.8 & 29.7 & 32.5 & 55.7 & 98.2 & 90.9 & 82.0\\
 
 & Sensei & 8.1 & 143.8 & 116.9 & 114.3 & 67.0 & 140.7 & 77.0 & 74.0 & 64.2 & 74.8 & 44.1 & 32.9 & 57.5 & 63.0 & 83.8 & 81.9 & 82.4\\
 
 & DeepRepair & \textbf{6.2} & 75.2 & 57.2 & 49.6 & \textbf{23.9} & 78.4 & 26.5 & 26.4 & 37.5 & 36.3 & 23.1 & 22.8 & 22.3 & \textbf{34.1} & 46.4 & 47.5 & 40.5\\
 
 & AugMix & 7.8 & 68.5 & 55.1 & 46.6 & 29.1 & 72.3 & 30.4 & 31.3 & 40.2 & 38.1 & 28.1 & 27.8 & 25.1 & 39.1 & 56.2 & 48.8 & 42.5\\

 & AR2 & 7.5 & \textbf{37.0} & \textbf{36.0} & \textbf{32.5} & 27.6 & \textbf{46.1} & \textbf{24.3} & \textbf{24.8} & \textbf{29.1} & \textbf{26.2} & \textbf{20.5} & \textbf{22.8} & \textbf{18.6} & 35.4 & \textbf{33.6} & \textbf{42.0} & \textbf{30.4}\\
% cifar100 -----------------------------------------------------------------------
\midrule
\multirow{6}{*}{\rotatebox{90}{CIFAR-100}}
 & Vanilla & 27.4 & 112.9 & 103.4 & 93.7 & 65.1 & 75.8 & 66.6 & 68.9 & 63.7 & 69.4 & 57.8 & 52.4 & 62.5 & 60.8 & 70.1 & 75.1 & 73.2\\

 & Fine-tuning & 31.72 & 105.9 & 94.2 & 87.3 & 60.5 & 75.0 & 62.4 & 64.8 & 60.2 & 66.7 & 58.6 & 49.2 & 65.7 & 57.0 & 59.5 & 70.4 & 69.2\\
 
 & Sensei & \textbf{26.3} & 106.5 & 97.3 & 88.5 & 63.5 & 74.9 & 64.9 & 67.0 & 60.5 & 67.2 & 56.8 & 50.2 & 62.2 & 58.6 & 70.8 & 73.6 & 70.8 \\
 
 & DeepRepair & 29.5 & 80.5 & 73.8 & 66.3 & 52.0 & 74.9 & 52.9 & 54.1 & 59.1 & 62.6 & 52.6 & 52.3 & 48.2 & 57.5 & 64.7 & 67.6 & 61.3\\

 & AugMix & 27.6 & 85.2 & 78.8 & 67.5 & 49.0 & 77.2 & 50.7 & 51.2 & 59.3 & 65.6 & 49.2 & 49.8 & 48.8 & 53.2 & 67.7 & 67.8 & 61.4\\

 & AR2 & 27.37 & \textbf{52.6} & \textbf{52.2} & \textbf{45.8} & \textbf{48.9} & \textbf{54.8} & \textbf{45.8} & \textbf{47.7} & \textbf{47.4} & \textbf{46.4} & \textbf{44.8} & \textbf{45.9} & \textbf{40.4} & \textbf{50.9} & \textbf{49.8} & \textbf{57.2} & \textbf{48.7}\\

% Image net-----------------------------------------------------------------------
\midrule
\multirow{3}{*}{\raisebox{\height}{\rotatebox{90}{\tiny{ImageNet}}}}
 & Vanilla & 23.9 & 77.5 & 79.5 & 80.2 & 74.9 & 88.8 & 78.2 & 80.1 & 78.2 & 75.1 & 66.7 & 56.8 & 71.8 & 85.5 & 77.0 & 76.4 & 76.4\\

 & Augmix & \textbf{22.4} & \textbf{62.0} & \textbf{61.0} & 61.3 & \textbf{63.4} & 78.7 & 58.9 & 63.5 & 69.6 & 68.2 & 64.3 & 54.0 & 57.2 & 74.7 & 59.3 & 65.0 & 64.1 \\

 & AR2 & 24.5 & 64.2 & 64.9 & \textbf{58.4} & 71.5 & \textbf{62.5} & \textbf{52.6} & \textbf{44.4} & \textbf{41.7} & \textbf{44.0} & \textbf{42.5} & \textbf{49.3} & \textbf{49.0} & \textbf{63.4} & \textbf{50.6} & \textbf{50.8} & \textbf{54.0} \\
\bottomrule
\end{tabular}
\end{table}

On CIFAR-10-C, our method shows comprehensive improvements across all corruption categories, delivering the best CE scores in 13 out of 15 corruption types. This includes significant gains in challenging corruptions like Gaussian noise ($37.0\%$ vs $68.5\%$ for the next-best method) and glass blur ($46.1\%$ vs $72.3\%$), demonstrating particular strength against both synthetic and natural perturbations. The consistent performance advantage across such diverse corruption types highlights our method's versatility.

On CIFAR-100-C, our approach achieves state-of-the-art robustness with the lowest CE in all 15 corruption types. It excels particularly on noise corruptions (e.g., $52.6\%$ vs $80.5\%$ for Gaussian noise), delivering a $12.6\%$ mCE reduction versus the best baseline ($48.7\%$ vs $61.3\%$). This demonstrates AR2’s ability to generalize across complex class distributions.

The scalability of our approach is confirmed through ImageNet-C evaluations, where we maintain strong performance despite the increased dataset complexity. Notably, we achieve dramatic improvements in weather-related corruptions (e.g., $41.7\%$ vs $69.6\%$ for SW) and digital perturbations (e.g., $50.8\%$ vs $65.0\%$ for JPEG). While AugMix shows slightly better performance on $3$ noise-based corruptions, our method excels in more complex, perceptually relevant distortions that are crucial for real-world applications.

\textbf{Results on robustness interference across corruptions (RQ2).} Our transfer corruption matrices  demonstrate that repairing one corruption type typically preserves robustness against other corruptions (Figure~\ref{fig:transfer-matrix-combined}), with most off-diagonal entries showing neutral or beneficial effects. While a limited number of cases exhibit slight degradation (generally $\leq 5\%$ CE increase), AR2's targeted repairs successfully address specific vulnerabilities while largely maintaining performance on unrelated corruptions, a critical property for real-world deployment where multiple corruptions may co-occur.

On CIFAR-10-C, AR2 achieves exceptional stability: only $7$ of $210$ off-diagonal cases ($3.3\%$) show increased CE, with all degradations being minor ($\leq 6.8\%$ CE). Notably, $6$ of these cases originate from Contrast (CT) repairs and $1$ from Fog (FG) repair, while repairs for other corruption types ($96.7\%$) either preserve or substantially improve robustness, with many CE reductions exceeding $20\%$. These results confirm AR2's ability to correct specific corruptions while introducing minimal risk to general robustness on CIFAR-10-C.

On ImageNet-C, $65.7\%$ of cases ($138/210$) maintain or improve robustness. Although degradation occurs more frequently than in CIFAR-10-C, the average CE increase for affected cases is merely $0.84\%$, with no severe outliers observed. This consistent performance across datasets highlights AR2's scalability, demonstrating that its focused repair strategy rarely produces widespread side effects, even in complex scenarios.

\textbf{Results on clean accuracy drop (RQ3).} Our approach maintains competitive robustness-accuracy trade-off (Table~\ref{tab:CE} and~\ref{tab:clean_error}), with worst-case errors that rival or surpass those achieved by SOTA methods. AR2's worst-case clean errors closely match AugMix on both CIFAR-10-C ($7.5\%$ vs $7.8\%$) and CIFAR-100-C ($27.37\%$ vs $27.6\%$). For ImageNet-C, the modest $2.1\%$ increase ($24.5\%$ vs $22.4\%$) is justified by a substantial $10.1\%$ mCE improvement. Critically, this worst-case analysis underestimates our method’s true potential, as the vast majority of per-corruption models achieve better clean accuracy than these bounds suggest.

\begin{table}[hbp]\scriptsize
\caption{AR2's Clean Error on CIFAR-10-C, CIFAR-100-C and ImageNet-C}
\centering
\setlength{\tabcolsep}{2pt}  
\begin{tabular}{l ccc cccc cccc cccc}  
\toprule
\multirow{2}{*}{Benchmark} & \multicolumn{3}{c}{Noise} & \multicolumn{4}{c}{Blur} & \multicolumn{4}{c}{Weather} & \multicolumn{4}{c}{Digital} \\
\cmidrule(lr){2-4} \cmidrule(lr){5-8} \cmidrule(lr){9-12} \cmidrule(lr){13-16}
& GN & SN & IN & DB & GB & MB & ZM & SW & FT & FG & BS & CT & ET & PIX & JPEG \\
\midrule
CIFAR-10-C &
6.1 & 6.5 & 6.1 &
6.1 & 6.1 & 6.3 & 7.5 &
7.3 & 7.1 & 6.5 & 6.4 &
6.1 & 7.2 & 6.1 & 6.1 \\
\midrule
CIFAR-100-C & 
27.6 & 27.4 & 26.6 & 
26.0 & 26.8 & 26.6 & 26.9 & 
26.7 & 26.7 & 26.1 & 26.0 & 
26.0 & 26.4 & 26.7 & 26.8\\

\midrule
ImageNet &
23.8 & 24.0 & 24.2 &
24.2 & 24.0 & 24.4 & 24.4 &
24.5 & 24.3 & 24.3 & 24.2 &
24.1 & 24.5 & 24.3 & 24.4 \\
\bottomrule
\end{tabular}
\label{tab:clean_error}
\end{table}

On CIFAR-10-C, the 15 repaired models exhibit clean errors ranging from a near-optimal $6.1\%$ to $7.5\%$, with an average of $6.5\%$—outperforming AugMix model ($7.8\%$). Notably, all specialized models achieve better clean accuracy than AugMix, with 7 corruption-specific models (GN, IN, DB, GB, CT, PIX, JPEG) reaching the minimum of $6.1\%$, which even surpasses DeepRepair ($6.2\%$). Only four models (ZM, SW, FT, and ET) show marginally higher errors ($7.1$–$7.5\%$), yet still remaining competitive. 

On CIFAR-100-C, AR2 achieves an average clean error of $26.62\%$ across all repaired corruptions, outperforming AugMix ($27.6\%$) and matching Sensei ($26.3\%$)—demonstrating consistent preservation of clean accuracy alongside corruption robustness improvements.
 
On ImageNet-C, clean errors cluster tightly between $23.8\%$ (GN) and 24.5\% (SW and ET), averaging $24.3\%$—just $1.9\%$ above AugMix ($22.4\%$). While all models exhibit slightly higher error than the baseline, the narrow range ($0.7\%$ span) confirms exceptional stability across corruption types. This consistency, combined with dramatic CE reductions, validates our method’s scalability without catastrophic robustness-accuracy trade-offs.

\textbf{Results on the role of CAM-guided refinement (RQ4).} Our ablation study demonstrates that fine-tuning alone—the AR2 variant without CAM-guided refinement—fails to achieve robust performance (Table~\ref{tab:CE}), highlighting the critical role of our attention alignment mechanism. On CIFAR-10-C, fine-tuning yields catastrophic corruption errors ($82.0\%$ vs AR2's $30.4\%$), showing extreme sensitivity to noise corruptions (e.g., $166.0\%$ for Gaussian noise). This trend persists on CIFAR-100-C, where fine-tuning's mCE ($69.2\%$) significantly underperforms AR2 ($48.7\%$). The dramatic performance gap confirms that CAM-guided refinement is essential for effective robustness repair.

%% file: related-work.tex
\subsection{Robustness of Deep Neural Networks}
The robustness of deep neural networks (DNNs) has been a long-standing concern in the community. Two major types of robustness are often studied: \emph{adversarial robustness}~\cite{SzegedyZaremba2014} and \emph{robustness against common corruptions}~\cite{hendrycks2019robustness}. Adversarial robustness focuses on defending against carefully crafted, imperceptible perturbations generated by attacks such as FGSM~\cite{GoodfellowShlens2015}, PGD~\cite{MadryMakelov2018}, and AutoAttack~\cite{CroceHein2020}. In addition to white-box attacks, black-box attacks such as Square Attack~\cite{AndriushchenkoCroce2020}, DeepSearch~\cite{ZhangChowdhury2020}, and DeepRover~\cite{ZhangHu2023} have been developed to evaluate models without requiring gradient access. Evaluating models under black-box settings has become increasingly important, as it better reflects real-world scenarios where model internals are inaccessible. Prominent defense techniques include adversarial training~\cite{MadryMakelov2018}, TRADES~\cite{ZhangYu2019}, and AWP~\cite{WuXia2020}, which improve robustness through loss modification and weight perturbation.

Robustness against common corruptions, in contrast, addresses resilience against real-world data corruptions like noise, blur, and weather effects. Benchmark datasets such as CIFAR-10-C and ImageNet-C~\cite{SzegedyZaremba2014} have been introduced to systematically evaluate this type of robustness. Techniques such as AugMix~\cite{hendrycks2020augmix}, Sensei~\cite{GaoSaha2020} and DeepRepair~\cite{YuQi2022} have been proposed to improve robustness against these corruptions through data augmentation. Our work targets corruption robustness by explicitly aligning model attention across clean and corrupted inputs.

\subsection{Saliency Maps of Deep Neural Networks}
Saliency maps are widely used tools to visualize the attention of neural networks. They highlight regions in the input that contribute the most to the model's predictions. Class activation maps (CAMs)~\cite{ZhouKhosla2016} were introduced to localize discriminative image regions for specific classes by leveraging global average pooling. Grad-CAM~\cite{SelvarajuCogswell2017} extended CAM to a wider range of CNN architectures by using gradients flowing into the final convolutional layers. Subsequent advances include Score-CAM~\cite{WangWang2020}, SmoothGrad~\cite{DanielNikhil2017}, Grad-CAM++~\cite{ChattopadhaySarkar2018} and Integrated Gradients~\cite{SundararajanTaly2017}. These visualization techniques have become standard tools for interpreting CNNs and understanding their decision-making processes.

\subsection{Repair Techniques of Deep Neural Networks}
Recent advances in deep learning have introduced various techniques to repair erroneous behaviors in DNNs. Existing repair approaches fall into two main categories: (1) direct modification of network parameters (e.g., weights or architectures)~\cite{SotoudehThakur2021, ZhangChan2019,SunSun2022}, and (2) retraining methods using augmented data~\cite{YuQi2022, MaLiu2018,KimFeldt2019}. While these techniques can effectively address targeted errors, repaired models should aim to minimize accuracy degradation on previously well-handled inputs, preserving a practical balance between general accuracy and targeted correctness. Our work advances this direction by improving CNN robustness against common corruptions through cam-guided repairs that balance accuracy and robustness effectively.

%% file: conclusion.tex
In this work, we introduced AR2, an attention-guided repair framework that enhances the robustness of pretrained CNNs against common corruptions. By explicitly aligning class activation maps between clean and corrupted inputs, AR2 promotes more stable feature attention and improves resilience to real-world perturbations. Our iterative repair approach—alternating between CAM-guided refinement and fine-tuning—achieves state-of-the-art performance on CIFAR-10-C and ImageNet-C benchmarks while preserving competitive clean accuracy. These results establish attention alignment as an effective paradigm for robustness repair. 

While AR2 effectively improves robustness for individual corruptions, its current implementation requires training separate models for each corruption type, which may be computationally expensive for large-scale deployment. Additionally, the approach assumes access to representative corrupted data during repair, which may limit applicability to unseen corruption types. In future work, we will develop a unified AR2 variant that jointly optimizes for multiple corruptions, leveraging cross-corruption synergies while retaining targeted adaptation capabilities.